\documentclass{article}

\usepackage[T1]{fontenc}
\usepackage[]{algorithm2e}

\usepackage{amsthm}
\usepackage{amsmath}
\usepackage{amssymb}
\usepackage{microtype}
\usepackage{graphicx}
\usepackage{caption}
\usepackage{subcaption}
\usepackage{booktabs} 

\usepackage{tikz}

\usepackage{hyperref}

\usepackage[accepted]{icml2021}

\icmltitlerunning{Updater-Extractor Architecture for Inductive World State Representations}

\begin{document}

\newcommand{\eps}{\varepsilon}
\newcommand{\extfunc}{f_{\varepsilon}}
\newcommand{\updfunc}{f_{\delta}}

\twocolumn[
\icmltitle{Updater-Extractor Architecture for \\
           Inductive World State Representations}

\icmlsetsymbol{equal}{*}

\begin{icmlauthorlist}
\icmlauthor{Arseny Moskvichev*}{ucicog}
\icmlauthor{James A. Liu*}{ksky}
\end{icmlauthorlist}

\icmlaffiliation{ucicog}{Department of Cognitive Sciences, University of California, Irvine, USA}
\icmlaffiliation{ksky}{K-Sky Limited, Hong Kong}

\icmlcorrespondingauthor{Arseny Moskvichev}{amoskvic@uci.edu}

\icmlkeywords{Continual Learning, Learning to Learn, Natural Language Processing, Zero Shot Learning, NLP, Truncated Backpropagation Through Time, TBPTT, Transformers}

\vskip 0.3in
]

\printAffiliationsAndNotice{\icmlEqualContribution} 

\begin{abstract}

Developing NLP models traditionally involves two stages - training and application. Retention of information acquired after training (at application time) is architecturally limited by the size of the model's context window (in the case of transformers), or by the practical difficulties associated with long sequences (in the case of RNNs). In this paper, we propose a novel transformer-based Updater-Extractor architecture and a training procedure that can work with sequences of arbitrary length and refine its knowledge about the world based on linguistic inputs. We explicitly train the model to incorporate incoming information into its world state representation, obtaining strong inductive generalization and the ability to handle extremely long-range dependencies. We prove a lemma that provides a theoretical basis for our approach. The result also provides insight into success and failure modes of models trained with variants of Truncated Back-Propagation Through Time (such as Transformer XL). Empirically, we investigate the model performance on three different tasks, demonstrating its promise. This preprint is still a work in progress. At present, we focused on easily interpretable tasks, leaving the application of the proposed ideas to practical NLP applications for the future.

\end{abstract}

\section{Introduction}

Language is a unique tool allowing us to learn new facts and adjust our beliefs and behaviour without having any first-hand experience. Reading a book or an article, as well as having a meaningful conversation with a friend may change our views on a number of moral, political, or professional issues. These views, in turn, directly affect our personal and professional decisions. Thus, in humans, processing natural language is tightly bound to the problem of adjusting one's beliefs about the world, and can have a lasting impact on one's behaviour. In other words, when humans process natural language, they continuously \emph{learn} from it.

In contrast, although modern language models achieve impressive results across a wide range of NLP applications \cite{AttentionIsAllYouNeed, LangmodelsAreUnsupMultitaskLearners}, they still \textbf{lack the ability to learn through language}. At application time, there is no mechanism that could result in a \textbf{long-term change in model's knowledge or behaviour based on linguistic inputs} that such a model receives. For example, a model can process a paragraph-sized text and perform the task it was trained for, such as question answering, summarization, named entity recognition, and so on. In some cases, a model can even generalize to new tasks in a zero-shot fashion \cite{LangmodelsAreUnsupMultitaskLearners}. And yet, it can not retain the knowledge between application-time instances. For example, summarising a Wikipedia article about World War II will have no effect on the model's ability to answer related questions in the future. For humans, it is not uncommon to recall or use a fact from a conversation they held many years ago. A typical model, in contrast, can only operate the facts that it implicitly learned in the course of training. There is currently no way to permanently incorporate new facts into its knowledge through natural language interaction alone (see further \autoref{sec:review}).

\paragraph{Why are there no models that can learn from language in the long term?} 

First, most models are unable to handle extremely long sequences. Therefore, models' interactions with the world are restricted to short-to-moderate length episodes, after which the model has to be ``reset'' to its default state, forgetting the interaction. A brute-force approach -- collapsing all model's ``experiences'' into one huge chunk of text and training the model on it -- is doomed to fail as computationally impractical. Second, there are not many datasets that require meaningful utilization of extremely long-term linguistic experiences.

In this paper, we propose a proof-of-concept transformer architecture with a consistent world state representation and a training procedure, addressing the first issue described above. We also demonstrate how a number of problems can be re-interpreted to fit the proposed framework, addressing the second issue. The theoretical part of our paper is devoted to studying the properties of the data distributions that are conducive to training models that can handle arbitrarily long input sequences. 

\textbf{The key idea behind our approach is to treat knowledge internalization as the main problem to focus on when developing general-purpose NLP models}. Our model maintains a consistent world state representation which can be updated and exhaustively queried as new information becomes available. This way, \textbf{we are explicitly training the model to change its beliefs about the world}, as opposed to the more traditional approach, where one hopes that the model will naturally infer a reasonable world state representation because the task requires it.

\subsection{Main contributions}\label{subsec:maincontrib}

\begin{itemize}
    \item Proposing a novel transformer-based architecture with a consistent world state representation.
    \item Identifying sufficient conditions on the data distribution under which cutting gradients between recurrent steps is justified both practically and theoretically.
    \item Demonstrating on two simulated tasks that the model has a number of desirable world-state-tracking properties, shows extremely strong generalization over episode lengths, and exhibits high tolerance to uncertainty and distribution shifts. 
    \item Showing that, when applied to the Pathfinder problem, a challenging benchmark, the model achieves competitive performance, while reducing memory requirements by an order of magnitude.
    
\end{itemize}

\section{Related work}\label{sec:review}

\subsection{Continual learning and Zero-Shot Learning in NLP}

The idea that AI models should learn ``on the fly'' (especially through verbal instructions) received a lot of attention in the early days of AI \cite{turing1950, BlockWorld}. Unfortunately, the complexity of modeling such learning in realistic scenarios via symbolic systems proved insurmountable, so the idea lost some of its popularity for a while. With the advent of deep learning architectures, however, many previously intractable problems became solvable. A number of related research directions have emerged to tackle various aspects of this idea: meta-learning, zero-shot learning, learning to learn, lifelong learning, and continual learning. 

Despite this, there has been relatively little cross-pollination between the fields of continual and zero-shot learning with language modeling itself. That is, language models are predominantly used as a backbone for zero-shot learning (see \citet{zslReview} for a review), but not as a target of zero-shot learning (although there are notable exceptions that use ZSL for specific NLP applications \cite{zeroShotLinking, hillEmbeddingTheDictionary, bengioOnTheFlyEmbs}). In other words, while many domains were able to utilize large-scale language models to model zero-shot ``learning from language'' in these domains, there has only been very scarce research in using purely verbal input for changing the knowledge of  the language model itself.

There are very few works that investigate general-purpose mechanisms one can use to adjust the transformer's knowledge at application time. The most notable contribution in this direction is \citet{explicitimplicit}, which demonstrated that pre-trained language models can be fine-tuned to incorporate new information ``on the fly''. The fundamental limitation of their approach, however, is that memory retention is strictly limited by the model's context window size.

\subsection{World state tracking in NLP}

\citet{weston2015babi} introduced the bAbI dataset, aimed to evaluate the ability for a model to quickly incorporate new data and build reasonable world state representations. One of the most successful models on that task is EntNet \cite{entnets}. While providing promising performance in many aspects, there a few limitations. First, the model is an extension of a traditional LSTM network, and, with the advent of transformer architectures, its practical value significantly diminished. Second, the model has to be trained with a variable number of timesteps in order to achieve generalization across the number of such timesteps, which contrasts sharply with our approach. Consequently, even though the long-range performance deteriorates slowly, it still does deteriorate enough to be detectable on the toy task considered in the paper, despite the efforts to expose the model to a variable number of sequence lengths.

In our work, we aim to demonstrate the benefits of employing a different approach, that of \emph{inductive world state representations}. In this approach, the model is trained extensively to incorporate a single set of facts into its representation, while retaining previous knowledge and accounting for knowledge ramifications.

That is, the model is trained to perform a single step of knowledge updating. If that skill is mastered, the task of incorporating arbitrary amounts of information becomes a matter of simply repeating the step as many times as needed. 

In that, our method is dramatically different from previous approaches in training recurrent networks. Usually, the underlying logic is that if the model is trained to work with sequences of lengths from 1 to n, one may hope that the model is going to naturally generalize to longer sequences. In our work, we don't need to rely on the network's natural generalization tendencies, since we \emph{directly} train the model on a task that is equivalent to working with sequences of arbitrary lengths.

\subsection{Handling longer sequences with transformers}

Recently, the need for models that can handle extremely long sequences started to become more and more obvious. A number of works investigated transformer modifications that can handle long sequences more efficiently (\citet{performer, linformer}, and many other works). Moreover, a number of benchmark challenges were proposed \cite{longrangearena} to allow for systematic evaluation of such models.

However, even for these models, their ability to handle long-range dependencies is fundamentally limited by their context window size as well as by memory limitations. While algorithmic and implementation optimization can yield fruitful results, substantially increasing the context window, it is still unclear whether it will ever be practical to use the context window to store lifetime-scale-long linguistic experiences. We explore an alternative recurrence-based approach, separating language processing and information internalization.

\citet{transformerxl} demonstrate that gradient-less recurrence can be a practical technique to expand the context window for language modeling tasks. On any step, their Transformer XL model re-uses the context vectors computed for the previous segment (without keeping the gradients), thus substantially increasing the maximal distance of information exchange. In contrast with our work, however, their mechanism does not allow for unlimited-horizon information exchange. More precisely, the maximal information-exchange distance is proportional to the product of the number of layers in the network times the size of the basic transformer context window. Our work builds upon their architecture, while our theoretical results suggest multiple pathways for further improvement.

\subsection{Backpropagation Through Time alternatives}

There have been a number of attempts to find alternatives to Backpropagation Through Time \cite{BPTT}, which is a standard (but memory-costly) and notoriously unstable way to train recurrent networks. A common modification is Truncated BPTT, which cuts the gradients after a number of steps. While Truncated BPTT sometimes leads to good results, it is also known to be biased and unstable \cite{mikolov2010recurrent}. \citet{tallec2017unbiasingTBPTT} propose a modification using varied truncation lengths and a re-weighting scheme which allows to alleviate the bias and instability, although still requiring more than one step of network unrolling (which is a substantial limitation for big models), and is marked by high variance. \cite{reviveRecBackprop} propose a completely different algorithm that allows to compute the gradients with constant per-timestep memory requirements, but imposes specific assumptions about the model architecture. Perhaps it is due to those restrictions that the proposed algorithms were not yet applied to transformer-based architectures.

The theoretical part of our work is substantially different in that we shift the focus to the data distribution and the sampling scheme. We demonstrate that focusing on just one step is enough, if the data distribution meets certain conditions. The practical part of our paper is completely different from the works mentioned in this section, since instead of training a traditional recurrent network, we introduce a novel transformer-based architecture and a training procedure allowing us to fully capitalize on our theoretical result.

\section{Problem setting}

We are interested in modeling the evolution of a world state, where two sources of information are present:

\begin{enumerate}
    \item World dynamics
    \item External instructions
\end{enumerate}

Intuitively, the first refers to changes that naturally follow from what the model knows about the world. For example, if the model is deployed as an NLP-based house assistant, it may know that every weekday kids go to school. Therefore, when the weekend is over, a model should update its world state representation, so that the question ``where are the kids'' results in an answer ``at school''.

On the other hand, certain pieces of information can not be predicted within the scope of what the model may know about the world. Continuing the example above, such a situation may arise if one of the kids gets a sore tooth and goes to the dentist instead of the school. A reasonable house assistant system needs to be able to meaningfully incorporate such information into the world state (which would imply changing the answers to queries like ``where are the kids'' and ``how much money do we owe to the insurance company'').

Other external world state updates may also reflect setting personal information (getting to know the family members, their tastes and preferences), changes in personal preferences (e.g. somebody wants to stop eating fast food), changes in one's occupation, moving to another house, and so on. While these events don't come out of nowhere and, on some level, may be predicted, their causes are out of the model's scope, and thus can be treated as arbitrary.

We will use this house assistant example throughout the paper to illustrate important points.

\subsection{Formal setup}\label{sec:formalsetup}

First, we need to introduce the notion of a \textbf{world state trajectory}. A \textbf{world state trajectory} $W$ (or simply \textbf{a world}) is an abstract entity with two important properties. First, it is indexed by time. Thus, $W_t$ represents a \textbf{world state} at time $t$ (time can be discrete or continuous, depending on the application). We will denote the space of all possible $W_t$ as $\mathcal{W}$.

Second, world states support the \textbf{information extraction} operation. Denote the space of all possible queries (statements about the world that may be true or not) as $\mathcal{Q}$. Then we have an \textbf{extractor function} $\extfunc$ defined on $\mathcal{W} \times \mathcal{Q}$. For any query $q \in \mathcal{Q}$, $\extfunc(W_t, q)$ represents a binary answer to this query (whether or not the query holds in the world $W$ at the moment of time $t$). When the query is provided along with its answer, it is an \textbf{instruction}, i.e. an \textbf{instruction} for a world $W$ at time $t$ is a pair $(\extfunc(W_t, q), q)$.

In the ``house assistant'' example, the instructions could be $I_0 = \{(True, \textrm{The house owner's name is John}),$ $(True, \textrm{John broke up with his girlfriend}) \}$, and $I_{1 \textrm{\footnotesize{year}}} = \{(True, \textrm{John is still single}) \}$. In this case the model should answer ``no'' to the query $q_{1 \textrm{\footnotesize{year}}} = (\textrm{John has a wife})$.

The notation is summarized in \autoref{tab:notation}.

We can now formulate the problem. At any time $t$, we must provide answers to all queries based on instructions that arrived during times $t': t' \leq t$. Formally, given a world $W$, at time $t'$ we receive a set of instructions $I_{t'} = \{ (\extfunc(W_{{t'}^{(k)}}, q^{(k)}), q^{(k)}) \}, k \in \{1 ... K\}$. We want to compute the value $\extfunc(W_t, q')$ for all $q' \in Q_t$ (all test queries at time $t$), using all instructions available at the time ($\{I_{t'}, t' \leq t\}$).

To make the model amenable to approximation, we assume that the worlds (world trajectories) and associated instructions come from some probability distribution $P_{W, I}$. We then define the \textbf{updater function} $\updfunc: \mathcal{W} \times \mathcal{I}^* \to \mathcal{P}(\mathcal{W})$ as $P(W_{t+1} | W_t, I_{t+1})$, where $\mathcal{P}(\mathcal{W})$ denotes the space of probability distributions over world states. In other words, the updater function outputs a distribution of world states at time $t + 1$, given a previous state and a set of incoming instructions at time $t$.

As further discussed in \autoref{section:modeldescription}, our model uses distributed representations (embeddings) $w_t$ for $W_t$ and approximates $\extfunc$ and $\updfunc$ with transformer modules \cite{AttentionIsAllYouNeed}. To initiate the process, $w_0$ is assumed to come from some fixed distribution. We use a point estimate to approximate the distribution $P(W_{t+1} | W_t, I_{t})$. We don't explicitly model the instruction probabilities (i.e. we condition on incoming instructions, but don't try to anticipate them). It is, however, a simple extension that may be relevant in certain applications.

\subsubsection{Narrowing the scope}\label{sec:narrowingthescope}

The problem described above is highly general. Many existing models can be interpreted in our notation. For example, traditional autoregressive language models can be interpreted as processing a single instruction (True, ``input word on position t is X'') on every step, and updating the world state representation accordingly. Predicting the next word from a hidden state is equivalent to providing an answer to all possible ``the word on position t+1 is $x$'' queries, where $x$ runs through all words in the vocabulary and only one query actually has ``yes'' as an answer. Time in this case runs from $1$ (the first word) to $n$ (sentence length).

In contrast, transformer language models \cite{AttentionIsAllYouNeed} can be seen as receiving all instructions and queries at a single time-step. There is no recurrent world state representation to update, so in our notation, there is only one processing time-step. The variable-length context representation is created from the set of incoming instructions (all in the form of ``the word in position k is x''), is used to answer all queries (e.g. ``the masked word in position n is x''), and then discarded.

In this paper, we focus on the problems with many processing steps (as in recurrent models), and many instructions and queries per step. The former ensures that we can work with sequences of arbitrary length (see further \autoref{subsec:dynstab}), the latter allows us to provide dense supervision on every step, training the model to properly update its beliefs about the world.

\begin{table}
\begin{tabular}{|p{40mm}|p{35mm}|}
\hline
\textbf{Abstract entity; Representation (if different)} & \textbf{Interpretation} \\ \hline
$W$ & World trajectory, World       \\ \hline
$W_t \in \mathcal{W}$; $w_t \in \mathbb{R}^n$& World state    \\ \hline
$q$      & A query        \\ \hline
$v=(0/1, q)$   & An instruction \\ \hline
$Q_t^{(i)} = \{q_k\}^{i}, Q_t^{(i)} \in \mathcal{Q}$ & Queries for the world $i$ at time $t$        \\ \hline
$I_t^{(i)} = \{v_k\}^{i}, I_t^{(i)} \in \mathcal{I} $   & Instructions for the world $i$ at time $t$        \\ \hline
$\extfunc: \mathcal{W} \times \mathcal{Q} \to \{0, 1\}$;       & Extractor function \\
$\hat{\extfunc}: W \times \mathcal{Q} \to [0, 1]$ &  

\\ \hline
$\updfunc: \mathcal{W} \times \mathcal{I} ^*  \to \mathcal{W}$;       & Updater function\\
$\hat{\updfunc}: W \times \mathcal{I}^*  \to W$ &
\\ \hline

\end{tabular}
\caption{Notation summary.
 It should be clear from context when we use a representation (e.g. query embedding versus an abstract query). The $\mathcal{I}^*$ notation denotes a sequence of instructions of finite length.}
 \label{tab:notation}
\end{table}

\section{Model description}\label{section:modeldescription}

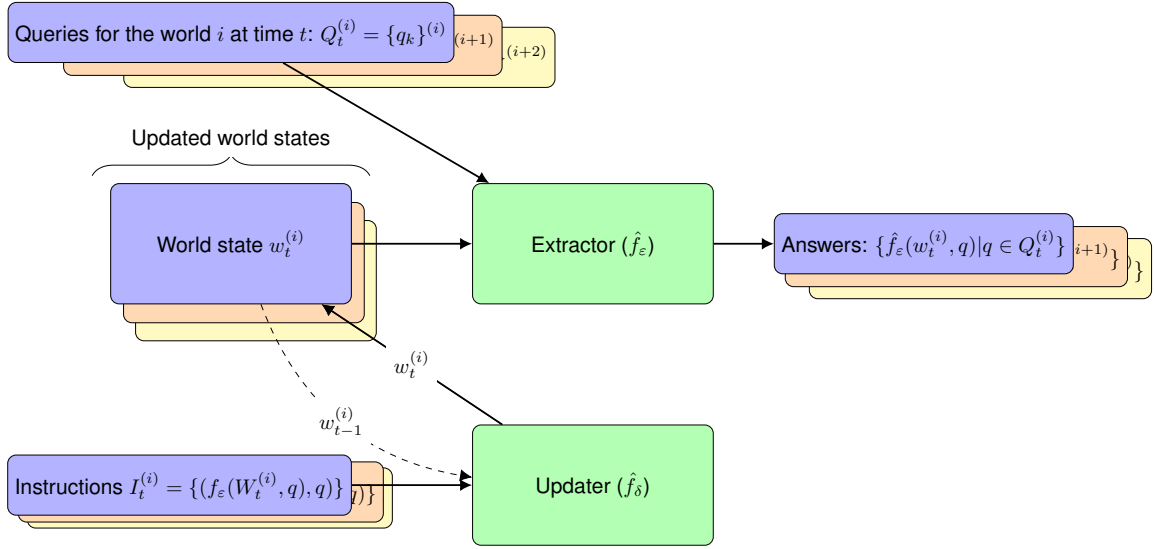
\begin{figure*}
    \centering
    \usetikzlibrary{shapes,arrows,shadows,positioning}
\usetikzlibrary{backgrounds}
\usetikzlibrary{decorations.pathreplacing}
\usetikzlibrary{arrows.meta}

\tikzset{%
  >={Latex[width=2mm,length=2mm]},
            base/.style = {rectangle, rounded corners, draw=black,
                           minimum width=4cm, minimum height=1cm,
                           text centered, font=\sffamily},
  memory/.style = {base, fill=blue!30, minimum height=2cm},
  memoryrep1/.style = {base, fill=yellow!30, minimum height=2cm},
  memoryrep2/.style = {base, fill=orange!30, minimum height=2cm},
  extractor/.style = {base, fill=green!30, minimum height=2cm},
  query/.style = {base, fill=blue!30, minimum height=1cm, minimum width=1.5cm},
  queryrep1/.style = {base, fill=yellow!30, minimum height=1cm, minimum width=1.5cm},
  queryrep2/.style = {base, fill=orange!30, minimum height=1cm, minimum width=1.5cm},
       startstop/.style = {base, fill=red!30},
    activityRuns/.style = {base, fill=green!30},
         process/.style = {base, minimum width=2.5cm, fill=orange!15,
                           font=\ttfamily},
}

\tikzstyle{note} = [rectangle, dashed, draw, fill=white, font=\footnotesize,
    text width=5em, text centered, rounded corners, minimum height=4em]

\scalebox{0.8}{

\begin{tikzpicture}[node distance=1.5cm,
    every node/.style={fill=white, font=\sffamily}, align=center]
  \node (memory)             [memory]              {World state $w_t^{(i)}$};
  \node (Extractor)             [extractor, right=20mm of memory]              {Extractor ($\hat{f}_{\varepsilon}$)};
  \node (Updater)             [extractor, below=20mm of Extractor]              {Updater ($\hat{f}_{\delta}$)};
  
  \node (Instructions)             [query, left=20mm of Updater]              {Instructions $I_t^{(i)} = \{(f_\varepsilon(W_t^{(i)}, q), q)\}$};
  
  \node (Answer)             [query, right=10mm of Extractor]              {Answers: $\{ \hat{f}_{\varepsilon}(w_t^{(i)}, q)| q \in Q_t^{(i)} \}$};
  
  \node (Query)             [query, above=20mm of memory]              {Queries for the world $i$ at time $t$: $Q^{(i)}_t= \{ q_k \}^{(i)}$};
   
   \node[above left=0.0 and 0.3cm of memory] (dummyl) {};
   \node[above right=0.0 and 0.3cm of memory] (dummyr) {};

  \begin{scope}[on background layer]
    \node [memoryrep1, below right=-14mm  and -36mm of memory, text width=2cm] (memrep1) {World state $\hat{W}_t^{(i+1)}$};
    \node [memoryrep2, below right=-17mm  and -38mm of memory, text width=2cm] (memrep2) {World state $\hat{W}_t^{(i+2)}$};
    
    \node [queryrep1, below right=-6mm  and -55mm of Query] (Queryrep1) {Queries for the world $i+1$ at time $t$ $\{ q_k \}^{(i+2)}$};
    \node [queryrep2, below right=-8mm  and -65mm of Query] (Queryrep2) {Queries for the world $i+1$ at time $t$: $\{ q_k \}^{(i+1)}$};
    
    \node [queryrep1, below right=-6mm  and -44mm of Answer] (Ansrep2) {Answers: $\{ f(W_t^{(i+2)}, q)| q \in Q_t^{(i+2)} \}$};
    
    \node [queryrep2, below right=-8mm  and -48mm of Answer] (Ansrep1) {Answers: $\{ f(W_t^{(i+1)}, q)| q \in Q_t^{(i+1)} \}$};

    \node [queryrep1, below right=-8mm  and -54mm of Instructions] (Insrep1) {Instructions $I_t^{(i)} = \{(f_\varepsilon(W_t^{(i+2)}, q), q)\}$};
    
    \node [queryrep2, below right=-9mm  and -55.5mm of Instructions] (Insrep2) {Instructions $I_t^{(i)} = \{(f_\varepsilon(W_t^{(i+1)}, q), q)\}$};
    

  \end{scope}
  
    
    \draw [decorate, decoration={brace,amplitude=10pt}]  
(dummyl)--(dummyr)  node [xshift=-0.0cm, yshift=0.6cm, midway] {Updated world states};
    \draw[->, line width=0.3mm] (memory) -- (Extractor);
    \draw[->, line width=0.3mm] (Query) -- (Extractor);
    \draw[->, line width=0.3mm] (Extractor) -- (Answer);
    \draw[->, line width=0.3mm] (Instructions) -- (Updater);
    \draw[->, line width=0.3mm] (Updater) -- (memory);
    
      

    \path
        (memory) edge[bend right, ->, dashed] node[midway] {$w_{t-1}^{(i)}$} (Updater);

    \path
        (Updater) edge[->] node[midway] {$w_{t}^{(i)}$} (memory);

    

  \end{tikzpicture}
  
}  
    \caption{Updater-Extractor Architecture.
    The notation is introduced in the \autoref{sec:formalsetup}.
    That dashed arrow indicates that no gradient is passed through the connection. The instructions and world state representation at time $t-1$ are passed to the updater which outputs a new world state representation for time $t$. This updated representation is then queried via the Extractor and the answers are compared to the ground truth.
    The gradient is \textbf{not} propagated through $w_{t-1}$, hence there is no need to store previous activations as in \cite{BPTT} or similar algorithms.
    }
    \label{fig:modeldescription}
\end{figure*}

The model structure is illustrated in \autoref{fig:modeldescription}.

The model performs two types of operations on world states: \emph{querying}, and \emph{updating}.

For \emph{querying}, the model receives a world state embedding $w_t$ and a query set $q$. Each query encodes a specific inquiry about the state of the world. For the house assistant example above, a query may encode a question ``Is John at home?''. The model $\hat{f}_{\varepsilon}(w_t, q)$ needs to approximate the true answer $\extfunc(W_t, q)$. The part of the model responsible for the query processing is called the \textbf{Extractor}. The Extractor is trained jointly with the updater, described below.

In the \emph{updating} operation, the model needs to process a set of instructions and incorporate them meaningfully into a world state representation $w_t$, obtaining $w_{t+1}$, while also accounting for the natural world state dynamics. This part of the model is called the \textbf{Updater}, as its role is analogous to the updater function $\updfunc$ introduced in \autoref{sec:formalsetup}.

The Updater training is mediated through the Extractor and the \emph{querying} operation. The world states that the Updater outputs must lead to correct query answers, when used by the Extractor. That is, if the Updater outputs $w_{t+1} = \hat{\updfunc}(w_t, I_t)$, we can obtain the Extractor predictions for a new set of queries, using $w_{t+1}$, as $\hat{Y}_{t+1} = \{ \hat{\extfunc}(w_{t+1}, q) | q \in Q \}$. These answers should approximate the true answers $Y=\{f(W_{t+1}, q) | q \in Q \}$.

\subsection{Architectural decisions}\label{sec:modelarchitecture}

\paragraph{Representing queries and instructions}

Each query and instruction is represented as a fixed sized vector (embedding). The way in which such embeddings are constructed is task-dependent. For example, if the task is easily interpreted in a knowledge-graph format, we can concatenate the embeddings of the source and target entities, together with a relation embedding (similar to works on Knowledge Graph embeddings, e.g. \cite{rescal}). In NLP applications, query embeddings may be represented by sentence or even paragraph embeddings obtained using a pre-trained language model, for example, BERT \cite{bert}.

\paragraph{Representing world states}

We split a fixed-length world state vector representation into a sequence of tokens with positional embeddings. This allows the world state to be passed into the Updater and Extractor directly, and lets the model to split information into different tokens.

\paragraph{Representing the Extractor and the Updater}

Both the Extractor and Updater modules are implemented as Transformers \cite{AttentionIsAllYouNeed}. We implement the Updater as a Transformer decoder that takes in the world state as input and uses the instruction tokens as context.
This way, the world state tokens use self-attention and cross-attention between the tokens and the instructions.
In contrast, we implement the Extractor as a decoder that takes in the queries as input and the world state as context. The decoder processes many queries in parallel, but we disable self-attention between queries (since queries are independent of each other).

\subsection{Relation to other models}\label{subsec:reltoothermodels}

The rationale for our modeling approach is best understood in comparison with existing methods, since in designing our model we aimed to address a number of specific limitations. In this section we discuss how our approach relates to the main two branches of sequence processing (recurrent models and transformer models).

\paragraph{Cutting the gradients}

Training recurrent networks is associated with a number of difficulties. The problem of exploding and/or vanishing gradients makes training difficult to handle in practice. Additionally, the necessity of storing unrolled network activations limits the sequence length due to memory constraints. Our model and training procedure address this problem in a radical way, cutting the gradients between different timesteps. Such an approach is an extreme case of Truncated BPTT. While usually Truncated BPTTs is seen as a rough approximation to BPTT, in \autoref{subsec:indtheory}, we demonstrate that under specific conditions, it is theoretically justified.

\paragraph{Opening the context window}

Transformers are architecturally limited in their ability to retain information. Transformers solve sequence processing problems using feed-forward activation with no recurrence, relying instead on attention mechanisms. This solves the problems mentioned in the previous paragraph, but introduces limitations of their own. While LSTMs and other recurrent networks (e.g. \cite{entnets}) allow (at least in principle) for infinite-horizon information storage and re-use, for transformers, the maximum range of information retention is limited by the size of the context window. It is impossible to arbitrarily increase the context window size, as it comes at a significant memory and computation cost. That is, when it comes to long-term memory, tranformers are architecturally limited.

In our model, we address this problem by adapting transformer models to be able to work with fully reusable world state representations.

\section{Inductive World State Representations}\label{sec:indreprs}

In this subsection, we provide the theoretical justification for cutting the gradients between different steps. We first lay out the intuition, and then formulate the technical result.

\subsection{Inductive World State Representations: intuition}\label{subsec:indintuitions}

\paragraph{Knowledge internalization eliminates the need for long gradient pathways.}

It is true that for any recurrent model, if there is a gap between when the information is introduced and when it is first used, then keeping the gradients flowing through the gap is necessary. However, if we can ensure that the model internalizes all relevant information into its world state representation, then the need for keeping gradients between steps becomes less pressing. We call such representations \textbf{inductive}.

Since an \textbf{inductive} representation contains all information that the model needs to perform the task, it becomes intuitively reasonable to train the model on individual steps from the sequence, as opposed to training on full sequences. This greatly reduces the memory requirements, and resolves the problem of exploding/vanishing gradients.

A natural way to ensure that the model incorporates all important information into the world state is to extensively query the world state representation on every step, making sure that no potentially useful information is lost. The architecture and the training procedure that we propose are designed to support such querying and to produce \textbf{inductive world state representations}. The next section formally demonstrates sufficient conditions on the training data distribution will result in obtaining inductive world state representations.

\subsection{Inductive World State Representations: theory}\label{subsec:indtheory}

In this section, we formalize the intuitions described in the previous section. Before introducing the result, however, we need a few additional definitions.

First, define a \emph{recall query} $q$ for an instruction $v_t$. A query $q_{t'}$ is a recall query for an instruction $v_{t}$ ($t \leq t'$) if $P(\{\extfunc(q, W_t)==\textrm{True}\}  | I_1 \dots I_t \setminus v_t \dots I_t') = 0$ and $P(\{\extfunc(q, W_t)==\textrm{True}\}  | I_1 \dots I_t \dots I_t') = 1$, for any instruction history s.t. $\exists t: v_t \in I_t$. In other words, the query is a recall query for a given instruction if the query answer is determined by whether or not the instruction was provided in the past. A simple real-life example could be a query $q_t=$ ``Did I not tell you to do your homework an hour ago?'', which is a recall query for an instruction $v_{t-1}=$ (``You must do your homework now'', True).

Next, we define a \emph{thorough} query distribution. We call a query distribution \emph{thorough} if for all instruction sequences $I$ with nonzero probability, for all times $t$, for all individual instructions $v_t$, $\forall t' \geq t$, there is a nonzero probability of sampling a recall query $q_{t'}$ for the instruction $v_t$. Intuitively, it means that all incoming information may turn out to be crucial at any point of time in the future.

Lastly, we need to define the notions of \textbf{stepwise-optimal} and \textbf{sequence-optimal} models. Consider a distribution of worls and instructions $P_{W, I}$. When we pass instructions $I_1 \dots I_{t-1}$ through a model, we also obtain a distribution over world state representations $P(w_t)$ at any time $t$.

A model is \textbf{stepwise-optimal} if $\forall W, I \sim P_{W, I}, \forall q_{t}$ such that the probability of sampling $q_t$ is positive, $P(q_{t}=True | w_{t}) = P(q_{t}=True | w_{t-1}, I_{t})$, where $w_{t} = \hat{\updfunc}(w_{t-1}, I_t)$.  That is, at every step, the model optimally uses all information passed through incoming instructions at time $t$ as well as any information potentially coming from previous steps through $w_{t-1}$.

A \textbf{sequence-optimal} model is a model such that $\forall W, I \sim P_{W, I}, \forall t > 0, \forall q_{t}$ such that the probability of sampling $q_t$ is positive, $P(q_t=True | w_t) = P(q_t=True | I_1 \dots I_{t})$. In other words, it is a globally optimal model that uses all incoming information and stores all relevant information in the world state representation.

Note that these two conditions only speak about the updater part of the model, as it is the most crucial part of the model responsible for incorporating incoming information and world dynamics into the world state. An optimal extractor is always, by definition $\hat{\extfunc}(w_t, q) = P(q=True | w_t)$

\newtheorem{theorem}{theorem}
\newtheorem{lemma}[theorem]{Lemma}

\begin{lemma}[on thorough querying]
\label{theor:thorquer}
Under thorough querying, any stepwise-optimal model is also sequence-optimal.
\end{lemma}

The proof is provided in supplementary materials (see \autoref{sec:thorquerproof}). The main idea is that since the model may have to re-use incoming instructions at any moment, the stepwise(locally)-optimal is incentivized to keep all relevant information in the world state representation. Then, since all useful information always remains available at any local step, a locally optimal model becomes globally optimal as well.

Note that the lemma does not hold without the thorough querying assumption. For example, if we provide one bit of information on step 1 and ask to recall it on step 3, with no queries using this information on steps 1-2, a model that completely ignores the first input and outputs a useless $w_t=0 \forall t$ is stepwise-optimal, but not sequence-optimal.

\paragraph{Theory consequences}

Lemma \ref{theor:thorquer} allows us to focus on making the model optimal on every single step (which can be done via regular gradient descent). The result guarantees that, as long as we organize a thorough training schedule, if we achieve stepwise-local optimality, the model will also be optimal globally. At present, we don't provide a full formal treatment of the behaviour of near-optimal models (which is what can be realistically obtained in practice), but we believe that it is reasonable to expect that the model would gradually come closer to global optimality as it approaches stepwise local optimality, as opposed to making an abrupt jump in global performance at the moment when true local optimality is achieved. This intuition is strongly supported by our experiments.

\section{Training procedure}

The pseudocode is provided in the \autoref{alg:trainingloop}. The procedure directly mirrors the theory. Multple queries on every step allow to add recall queries when needed, thus satisfying the \emph{thorough querying} condition. Because of that, we can safely cut the gradients between different steps. At the same time, there are a few subtleties. Firstly, the purist way to implement gradient descent would be to sample independent world state transitions $w_t = \hat{\updfunc}(w_{t-1}, I_t)$. In practice,  obtaining $w_{t-1}$ still requires rolling the network forward. To avoid wasting representations computed in the process, we accumulate the individual step gradients along the sampled trajectories, even though it makes the samples in the batch correlated. Consequently, it is possible to apply gradients at different times during the trajectory. For example, one may accumulate gradients over all timesteps before making a single gradient step, or, on the other extreme, one may make a gradient step after every single instruction-query step. We treat this update frequency as a hyperparameter.

\begin{algorithm}[H]
 \For{ N outer cycles }{
  \KwData{sample K world trajectories $W^{(1)} , \dots, W^{(K)}$, each of length $T$}
  Initialize K world state representations $w_0^{(1)}, w_0^{(2)} ... w_0^{(K)}$ (randomly or by setting all equal to 0).\;
  
  \For{$t$ in $1 \dots T$}{
  
     \For{$k$ in $1 \dots K$}{
     Sample \emph{instructions} $I_t^{(k)}$ valid at time $t$.
     
    Obtain new world state representations

    $w^k_{t} = \hat{\updfunc}(STOP\_GRADIENT(w_{t-1}^k), I_t)$.

     Sample \emph{queries} $Q_t^k$, obtain model predictions
     $\hat{Y} = \hat{\extfunc}(w_{t}^{(k)}, Q_t^k)$.
     
     Compute the loss
     $L(\hat{Y}, \extfunc(W_{t}^{(k)}, Q_t^k))$.
     
    Backpropagate the loss gradients.
     
     }
     
     \uIf{t \% update\_freq = 0}{

        Make a gradient step, zero accumulated gradients
     }
  }
  
 }
 \caption{Training procedure pseudocode}
 \label{alg:trainingloop}
     
\end{algorithm}

\section{Experiments}

First, we directly test the theoretical results on a simple LSTM network. We then evaluate our full model architecture on three synthetic tasks. The first two of them demonstrate some of the qualitative benefits of representations obtained using our training procedure and architecture. The last experiment demonstrates competitive performance on a benchmark that is known to be challenging for transformers.

\section{Experiment 0: LSTM recall}

In this experiment, we directly tested the theory developed in \autoref{subsec:indtheory} on a single-layer LSTM architecture \cite{LSTM}. We first describe an insightful failure case that strongly violates the thorough querying condition described in Lemma \ref{theor:thorquer}. Then we describe two successful scenarios, one of which exactly satisfies the thorough query condition, while the other satisfies its weaker version, although still resulting in optimal performance.

\subsection{Failure case: thorough querying violation}

The task is extremely simple. We have a vocabulary of $K$ distinct tokens (one of them is a special RECALL token). The model receives a sequence of tokens of length $T$. The token provided on the initial step is the \emph{target token} for the sequence. On all steps, if the input is any token other than RECALL, the model must repeat the input in its answer. Otherwise, the model must output the token memorized on the first step. The last token was always the RECALL. For example, for an input sequence ``4, 3, 6, RECALL, 3, RECALL'', the correct output sequence is ``4, 3, 6, 4, 3, 4''. We used 10 different tokens and a sequence length of 12. We randomly generated a sequence of numbers and then flipped input tokens in every position (except the first) to RECALL with a probability of 0.3. 

In this setting, the updater is represented by all parts of the LSTM that update the hidden state (using the previous hidden state and the input token). The extractor is a simple linear layer mapping the hidden state to a distribution (logits) over output tokens.

The task is, of course, easily learned by an LSTM model with full BPTT training \footnote{We trained the model using AdamW \cite{adamw} with a batch size of 128, learning rate of 1e-4 and default parameters otherwise. We used a standard cross-entropy loss function. The LSTM had the hidden state dimension of 64.}, but cutting the gradients between different steps results in a complete failure. When we cut the gradients, the model learns to copy the inputs, but fails the recall task. At the last step (which is always a RECALL step), the model performs at chance.

The reason it happens is that the thorough querying condition is not satisfied. When the model receives a token other than RECALL on any step, there is no (local) incentive for the model to not forget the information about the hidden token. In traditional sequence-to-sequence modeling, the set of queries is exactly the same on every step: \{$y_t = A?$, $y_t=B?$, $y_t=C?$... \}. Because of that, for an instruction sequence $4, 6$, after the instruction $6$ is provided, there is no query left which answer would depend on the fact that the target token is $4$.

It illustrates the crucial difference between traditional sequence-to-sequence training and our approach. In sequence-to-sequence training, the set of queries is rigid: the same queries are answered on every step, and if a certain piece of information is not immediately relevant given the incoming instructions (the input is not requesting it) - there is no local incentive to keep it in the world state representation. Unless the representation is of infinite capacity, non-thorough querying actively incentivizes forgetting past information.

This negative result highlights another limitation of traditional sequence to sequence training: the boundary between the instructions and queries is blurred. The inputs are used to both provide information about the world and to request the necessary information out of the network (like in our recall task). The absence of independent mechanisms for probing the world state makes it difficult to investigate the model's beliefs about the world or to explicitly train it to change those beliefs. That is, the information about the consequences of any incoming information must be spoon-fed, one element at the time, as there is usually only one answer at any timestep.

\subsection{Thorough querying success cases}

In order to fully satisfy the \emph{thorough querying} requirement, we need to disentangle providing instructions and querying, which requires a slight modification to the architecture and the data itself (there is no reason to have RECALL inputs anymore). The hidden states that are generated are not used to generate just one answer, but are rather either queried about the target token, or about the current incoming token. To allow for independent querying, instead of a linear mapping from the hidden state directly to output tokens, we first concatenate the hidden state with the query encoding. There are only two different queries: ``recall'' and ''repeat''. Since in this case the updater does not know what the hidden state is going to be queried about, it remains optimal to remember the target token (apart from encoding the current input). This setting fully satisfies the \emph{thorough querying} assumption. Consequently, the model quickly reaches ceiling performance.

The second way to achieve ceiling performance is by slightly modifying the training data distribution, while keeping the original LSTM architecture. It can be achieved by introducing \emph{reminder noise} into the data.  At any step, with a fixed probability (we used 0.05), the correct answer is replaced with the target token for the given sequence. While it does not strictly satisfy the conditions of Lemma \ref{theor:thorquer}, it does make it beneficial for the model to keep at least some probability allocated to the target token (because we use cross-entropy loss), even if the input does not request it. Therefore, it is never beneficial to completely forget the target token. As a result, even though the data becomes noisier, the model reaches ceiling performance (accuracy of 1 if we remove reminder noise at test time). This result suggests that the sufficient conditions in Lemma \ref{theor:thorquer} can be made weaker.

Lastly, as an additional test, we varied the time at which the first recall query (for the modified LSTM case) or the reminder noise (for the second approach) appeared in the dataset. As expected, we observe that performance drops substantially as the location of the first recall query is shifted away from the beginning of the sequence. This happens because there is no incentive to maintain the target token information during the gap steps (the steps with no recall queries or reminder noise). Surprisingly, however, the model still performs above chance, although learning much slower, with non-monotonic jumps in the loss function, and usually not converging on the optimal solution. This behaviour suggests that the model manages to pick up the traces of the initially provided information, even if the sufficient conditions in Lemma \ref{theor:thorquer} are not satistied.

\subsection{Experiment 0: result discussion}

We observe that as long as the thorough querying condition is satisfied, truncated gradient training results in a globally optimal model. These results provide direct empirical support to the Lemma \autoref{theor:thorquer}. Moreover, it turns out that even partial satisfaction of the thorough querying condition sometimes allows to achieve ceiling performance. Instead of having strict recall queries (queries that are fully determined by the event $v_t \in I$), it may sometimes suffice to have queries that \emph{change} their probability depending on the event $v_t \in I$. Similarly, the model manages to achieve above-chance performance in recall queries even if there are steps on which there are no recall queries and the Lemma conditions are not satisfied. Overall, however, when the conditions of Lemma \ref{theor:thorquer} are not met, the training is unstable and often results in suboptimal models, which accords with existing literature on Truncated BPTT \cite{mikolov2010recurrent, tallec2017unbiasingTBPTT}.

Overall, the results support our theoretical development, while also pointing out that there is leeway around the sufficient conditions. In other words, the results suggest that lemma conditions, while sufficient, are not necessary, as they are too restrictive. We believe that this motivates further theoretical development of the topic, since obtaining weaker sufficient conditions may have substantial practical implications.

\section{Experiment 1: World of Numbers}

The first experiment using our transformer-based architecture is the ``world of numbers'' problem. In this setting, the world states are n-tuples of handwritten digit images. The queries are in the form $(e_1, e_2, r)$, where $e_1$ and $e_2$ represent pixel coordinates and $r$ represents the index of the image. For example, the query $(14, 7, 3)$, requests the pixel at $x=14$, $y=7$ from the third image in the n-tuple. We use a binarized version of the MNIST dataset \cite{MNISTlecun} to obtain the images.

\begin{figure}
    \centering
    \begin{subfigure}{.5\linewidth}
      \centering
      \includegraphics[width=.7\linewidth]{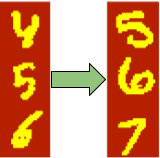}
      \caption{}
      \label{fig:wstrans}
    \end{subfigure}%
    \begin{subfigure}{.5\linewidth}
      \centering
      \includegraphics[width=.8\linewidth]{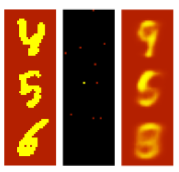}
      \caption{}
      \label{fig:sparsinf}
    \end{subfigure}
    \caption{``World of numbers'' problem setting.
    \footnotesize{a) World state transition example.  b) Providing sparse information. Left column - true world. Middle - information given at step 1 (black pixels are not shown to the model). Right - the model's predictions about the world after only receiving the information in the middle column. Since very little information was given, the model struggles and predicts generic shapes.}}
    \label{fig:worldofnumbers}

\end{figure}

The world dynamics are simple: at each timestep, the n-tuple is ``semantically'' rotated forward. That is, if the initial tuple at $t_0$ comprises images of digits 1, 2, 3, 4, and 5, the tuple on the step $t_1$ will comprise the (newly sampled) images of digits 2, 3, 4, 5, and 6 (see \autoref{fig:worldofnumbers}).

On every step, the model is provided with information about some of the pixel values and is queried about the values of other pixels. To make the task more challenging, we only provide very limited: we sample between 0 and 75 pixels out of 2352 (i.e. $28^2 \cdot 3$) as instructions on every step. That is, the data that the model receives is very sparse, so it has to rely on its knowledge of the world dynamics.

The goal of this simple experiment is to investigate the model's capacity to follow instructions, learn global world dynamics, and, lastly, its ability to handle uncertainty.

\subsection{Learning world dynamics \& evaluating trajectory stability}\label{subsec:dynstab}

First, we test the model's ability to learn the world dynamics. We train the model by providing values of 0-500 pixels at $t=1$, and then providing additional values of 0-75 pixels on every step thereafter. Additionally, on every step, the model is queried about the value of various pixels, both observed and unobserved (75 pixels total).

We advance the worlds eight time steps forward during training and we don't propagate gradients between different steps.

The first important result is that the model quickly reaches ceiling performance when it comes to incorporating pixel value information given directly (i.e. if a pixel value was provided at time t, the model is able to remember that value). That means that the updater and extractor learn to meaningfully coordinate their actions, using world state representations to store/communicate all relevant information.

\paragraph{Learning semantic world dynamics}

More importantly, the model learns to predict appropriate values for unobserved pixels, extrapolating from the available information. The model also learns the semantics of the world dynamics. That is, even if no information is provided during some steps (at application time), the model appropriately keeps track of the state of the world. For example, if the first step provided ample information about images of digits (5, 6, 7), the model infers that the third step must contain pictures of digits (7, 8, 9) and predicts appropriate generic shapes, even though it has not information about specific instances of digits in question. 

\begin{figure}
    \centering
    \begin{subfigure}{.5\linewidth}
      \centering
      \includegraphics[width=.7\linewidth]{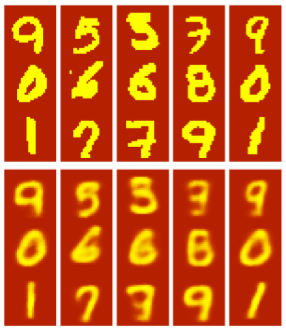}
      \caption{}
      \label{fig:afterinig}
    \end{subfigure}%
    \begin{subfigure}{.5\linewidth}
      \centering
      \includegraphics[width=.7\linewidth]{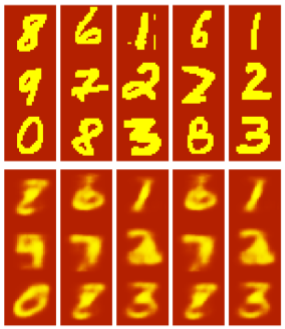}
      \caption{}
      \label{fig:roll100}
    \end{subfigure}
    \caption{Trajectory stability. \footnotesize{a) Top-true world state on step 1. Bottom-model beliefs on step 1, after 1500 out of 2352 pixels values are provided. Notice that the model captures each handwritten digit style: for example, ones are tilted at a different angle, reflecting the data. b) A world at t=50, where again, all information is given on the first step, with no input afterwards. The model has no information about specific digit instances, but knows (from step 1) about their identities. Therefore, the model predicts generic digits. Notice that all digits having the same identity are reconstructed identically. Notably, rolling a world, 10, 1000 or a 10000 step forward with no input information results in visually identical reconstruction, showing that the model retains its world state knowledge across apparently arbitrary horizons.}}
    \label{fig:rollingworlds}

\end{figure}

\paragraph{Trajectory stability and inductive generalization}

Although the model is only trained for up to eight time steps, its performance does not degrade over time, even when tested for many more time steps (see \autoref{fig:rollingworlds}). This again strongly demonstrates that the model learned generic rules for semantic world state advancement, as opposed to learning to make a specific number of world state adjustments, as it often happens with LSTMs.

\paragraph{Handling sparse information}

The model also has an almost uncanny ability to glean information from extremely sparse data. When initialized from a random world state, by receiving just five pixels per step, the model manages to infer the general shapes of the digits after a few steps. It takes substantially more steps, however, for the model to form a correct belief about the underlying semantic state, compared to the situation when a larger number of pixels is provided. This shows that the model is essentially integrating incoming information that is provided in very scarce ``doses'' over time, while also being able to adjust its beliefs faster, if more evidence is provided at once. Such behaviour is, of course, highly desirable for any AI system.

\section{Experiment 2: Game of Life with interventions}

The ``world of numbers'' experiment handles worlds with a relatively small number of possible ``semantic'' world states (even though the actual images may vary greatly, in total, there are only 10 different image classes). 

In this experiment, we test the model's ability to handle much more complex scenarios, where the space of possible world states is exponentially large and requires modeling local entity-to-entity interactions between timesteps.

For this purpose, we created a \textbf{Game of Life with interventions} environment. In this experiment, an 8 by 8 grid world evolves according to the rules of Conway's Game of Life (see \citet{gameoflife}). Queries are in the form $(x, y)$ requesting information about the state (dead or alive) of the cell with corresponding coordinates at time $t$.

To test whether the model can handle arbitrary external changes in the state of the world, we introduce the notion of an intervention: an arbitrary (not predictable from world dynamics) instruction to change the state of any specific cell. The model should be able to instantly incorporate these \textbf{interventions} into the world state and meaningfully propagate them forward in time.

\subsection{Experiment 2: result discussion}

We successfully train two models, one on worlds without interventions and on worlds with variable numbers of interventions on each time step. In both cases, we provide all pixels at the first time step. In the worlds without interventions, we do not provide any inputs in subsequent steps, and in the world with interventions, we provide inputs corresponding to these interventions.

As before, the models are able to continue correctly updating the model over an arbitrary number of steps during test time (we tested the model by rolling the world up to 10000 steps, with no drop in performance).

To put these results in perspective, it may be useful to compare them to the toy dataset results in \cite{entnets}. In that paper, the authors track a world state on a 10x10 grid with 2 agents, each of which have 4 possible states (facing top/down/left/right) and can be located in any square. The agents also don't affect each other in any way. This world has $16 \cdot 10^4 < 2^8$ possible states. 

In contrast, the game of life on an 8x8 grid has $2^{64}$ possible initial states and involves learning non-trivial agent interaction dynamics. In addition, interventions pose an additional challenge as they require the model to break the natural world dynamics upon request.

The problem dynamics considered in \cite{entnets} can be exhaustively learned even by a very moderately sized network. In the Game of Life with interventions, exhaustively memorizing $2^{64}$ state trajectories is not feasible, so the model has to internalize the rules in order to perform well.

\section{Experiment 3: Progressive Pathfinder}

\begin{figure}
    \centering
    \includegraphics[scale=.9]{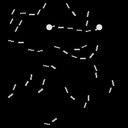}
    \includegraphics[scale=.9]{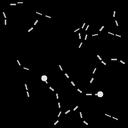}
    \caption{Pathfinder challenge problem example. The task is to determine whether the big dots lie on the same path (left) or on different paths (right). This task proved to be fairly challenging for a number of models \cite{pathfinder2010, longrangearena}.
    \label{fig:pathfinderProblemSetting}}
\end{figure}

To test the model in more challenging setting, we applied our model to the Pathfinder problem \cite{pathfinder2010}. This task was re-introduced as a machine learning benchmark by \citet{linsley2019learning}, who found that CNNs were inefficient at learning the requisite long-range dependencies to solve the problem. \cite{longrangearena} use the Pathfinder problem as one of the benchmark tasks for testing transformers and their ability to handle long range information(see \autoref{fig:pathfinderProblemSetting}).

To re-interpret the static problem in order to fit our sequential setting, we randomly group the pixels of an instance into a number of equally-sized partitions. At each time step, we feed a new partition to the model, and the model is then asked to provide the values of previously observed pixels, as well as to predict the class of the instance. In order to solve the problem, the model must find a world state representation that is flexible enough to handle the reception of information in an unordered manner, yet structured enough to readily decode the relevant long-range path information in order to correctly classify the instance.

In this setting, the model functions similarly to PixelCNN \cite{oord2016conditional} (although note that PixelCNN is not generally used for classification). However, traditional PixelCNN models require fixed-order inputs and are difficult to train due to their sequential nature. We are able to take advantage of the parallelism inherent in transformer-based architectures to avoid these drawbacks.

In comparison to the more standard transformer models tested in \citet{longrangearena}, our model is more memory-efficient as it does not need to receive the entire image at once. Furthermore, using a vanilla transformer architecture, we reach a competitive accuracy (0.73), which is better than most specialized long-range architectures reported in \cite{longrangearena}. This demonstrates that mixing parallelism and recurrence can be more effective than simply trying to use an extremely large context window.

\section{General Discussion}

We have demonstrated that by using Inductive World State Representations we can adapt transformers to work with sequences of arbitrary length. We obtain this ambitious result by explicitly training the model to incorporate incoming information into its world state representation.  Our procedure allows to directly train the model to make meaningful generalizations based on available data.

Moreover, the model does not have to generalize over variable sequence lengths in the same sense in which previous recurrent models had to. Instead of exposing the model to various sequence lengths in the hopes that the model will naturally infer the pattern and generalize its behaviour (a task in which recurrent models fail in unexpected and surprising ways \cite{lake2018still}), we train our model to make a single knowledge update step, but to make it as precisely as possible. Generalization is then supported by simple inductive logic: internalizing information from N chunks of input data can be achieved by learning all you can from each chunk and processing them one by one.

The model learns to utilize intrinsic world dynamics and to meaningfully incorporate external updates into its world state representation. This makes is suitable for a wide range of applications. The most promising direction, in our view, is that of continuously-learning NLP agents (such as personal assistants), which naturally involve arbitrary-length sequences.

Another crucial point of our paper is that it is often reasonable to drop the assumption that the data distribution is something that is out of the researcher's/practitioner's control. With the widespread use of transfer learning, for many models, most important applications are outside of the original task they were trained on. We believe that it is reasonable to focus on determining the features that the model should have, and on what are the crucial properties of the data that would help the model to learn these desirable features. The theoretical part of our paper is devoted to studying the properties of the data distributions that are conducive to training models that can meaningfully handle and learn ``on the fly'' from arbitrary long input sequences.

There are two limitations of our approach. Firstly, the data for our model must be structured differently than in traditional end-to-end training. In some cases (as in the Pathfinder problem), simple re-interpretation is sufficient and leads to good performance, but in other cases (e.g. bAbI \cite{weston2015babi}) the datasets in their original form may become unusable.

Another limitation is that for the tasks that require rote memorization (such as Game of Life, where knowing one fact (pixel activation) tells very little about whether or not other facts are true), a very high knowledge retention rate is required. Otherwise, the model knowledge will quickly deteriorate. For example, even if the model forgets only 1\% of facts on every step, in 250 steps, it will only retain 10 \% of its original knowledge. This was not an issue for Game of Life with Interventions and Pathfinder tasks but it may have been one of the reasons why the model struggled with Pathfinder XL.

This limitation is less dramatic for the domains allowing for rich ``common sense'' reasoning, i.e. where different pieces of information are highly entangled. Such is the case with World of Numbers task. In this situation, the world structure itself allows to ``triangulate'' forgotten facts based on retained information, thus maintaining a coherent general picture. In this case, forgetting a value of any specific pixel bears very little impact on the model performance, since the model remembers the general structure of the world.

This behaviour is highly realistic from a cognitive standpoint. It is well known that humans are not very adept in remembering isolated facts in an unfamiliar context. At the same time, it is common knowledge that a specialist may remember enormous amounts of information in their domain. It is not just a consequence of impeccable memory - even the best specialists in their field are not immune to forgetting (see, for example, R.P. Feynman's  discussion on the impossibility of simply memorizing formulas \cite{feynman2011feynman}). A professional, however, can reconstruct (``triangulate'') any forgotten knowledge based on related facts that they still remember.

Similarly, people have enormous amounts of common-sense knowledge, and the ease with which we operate with it may be explained by the fact that most commons-sense facts are highly entangled. For example, if somebody were to forget the name of their mother, they may recover this knowledge by recalling a situation in which somebody called their mother by her name. If somebody were to forget that cats are mammals, one can reconstruct this knowledge by thinking of other similar domestic animals. As we demonstrated in the World of Numbers experiment, our model naturally learns such triangulation behaviour, allowing it to retain general information indefinitely, although not being immune to forgetting specific low-level details.

Overall, the proposed approach offers a different perspective to problems involving lifelong language-based learning and world-state tracking. We hope that in the future, more applications of our method will be explored.

\emph{This is still a work in progress, and we welcome any feedback.}

\bibliography{main}
\bibliographystyle{icml2021}

\section*{Appendix A: Thorough querying lemma proof}

\label{sec:thorquerproof}

\subsection*{Assumptions, notation recap, and the lemma statement}

We assume that there is a probability distribution over world trajectories, associated instructions, and queries $W, I, Q \sim P_{W, I, Q}$. Sampled instructions $I$ contain instruction sets for every step, i.e. $\{I_1, I_2, I_3 \dots I_n \dots \}$.  We assume that the space of all possible instruction sequences is countable (and therefore we must assume that we are working with sequences of finite length or, for convenience, infinite sequences with redundant (repeating) tails).

\textbf{Notation details for the proof}. For convenience, we denote $\{I_1 ... I_n \}$ as $I_{1 \dots n}$. Additionally, we use $I_{1...n} \setminus v_t$ to denote $\{I_1, \dots, I_t \setminus v_t, \dots, I_n \}$. $Q$ contains queries sampled for every step. For simplicity, we assume that we only sample individual queries (as opposed to query sets) for every step. That is, $Q = \{q_1, q_2, q_3, \dots \}$. In the proof, we use $P(q=True | I_{1 \dots n})$ or $P(q | I_{1 \dots n})$ instead of $P(\{\extfunc(q, W_t)=\textrm{True}\} | I_{1 \dots n})$.

A query $q_{t'}$ is a \emph{recall query} for an instruction $v_{t}$ ($t \leq t'$) if $P(\{\extfunc(W_t, q_{t'})=\textrm{True}\}  | I_1 \dots I_t \setminus v_t \dots I_t') = 0$ and $P(\{\extfunc(W_t, q_{t'})=\textrm{True}\}  | I_1 \dots I_t \dots I_t') = 1$, for any instruction history s.t. $\exists t: v_t \in I_t$.

A query distribution is called \emph{thorough} if for all instruction sequences $I$ with nonzero probability, for all times $t$, for all individual instructions $v_t$, $\forall t' \geq t$, there is a nonzero probability of sampling a recall query $q_{t'}$ for the instruction $v_t$. Intuitively, it means that all incoming information may turn out to be crucial at any point of time in the future.

Next, fix a model (a pair $\hat{\updfunc}, \hat{\extfunc}$).
We define random variables $w_{t}= \hat{\updfunc}(w_{t-1}, I_t)$ for all $t>0$.
We assume that $w_0$ is either a constant or a random variable sampled independently of $I$, $W$ and $Q$. We also assume the $\hat{\updfunc}, \hat{\extfunc}$ output point estimates and are deterministic.

A model is \textbf{stepwise-optimal} if $\forall W, I \sim P_{W, I}, \forall t > 0, \forall q_{t}$ (such that the probability of sampling $q_t$ is positive), $P(q_{t}=True | w_{t}) = P(q_{t}=True | w_{t-1}, I_{t})$.

A \textbf{sequence-optimal} model is a model such that $\forall W, I \sim P_{W, I}, \forall t > 0, \forall q_{t}$ (such that the probability of sampling $q_t$ is positive), $P(q_t=True | w_t) = P(q_t=True | I_1 \dots I_{t})$.

Lastly, the statement of Lemma \ref{theor:thorquer} is as follows: \emph{under thorough querying, any stepwise-optimal model is also sequence-optimal}.

\subsection{The proof}

Assume that the model $\hat{\updfunc}$ is stepwise-optimal. Then the model is also sequence-optimal, by induction.

\paragraph{Base}

Note that $w_0$ is a world state representation before any information is provided. It is either a fixed constant for the initial world state representation or a random variable drawn from some fixed initialization distribution. That is, $w_0$ is independent from $W, I$, and $Q$. 

Therefore, at $t=1, \forall q_1, p(q_1=True | I_1, w_0) = p(q_1=True | I_1)$. Then, by stepwise-optimality, we have $ P(q_1=True | w_1) = P(q_1=True | I_1)$, which is the condition for sequence optimality at $t=1$.

\paragraph{Step}

Assume that sequence optimality holds for all $t \leq n \in \mathbb{N}$. We want to show that at $t = n + 1$, the condition holds as well, i.e. that $\forall W, I \sim P_{W, I}$, $\forall q_{n+1}$ with positive sampling probability, $P(q_{n+1} | w_{n}, I_{n+1}) = P(q_{n+1}=True | I_{1, \dots n+1})$.

Let's consider an arbitrary $q_{n+1}$ with a nonzero sampling probability.

First, note that by stepwise-optimality, we have $P(q_{n+1}=True | w_{n+1}) = P(q_{n+1}=True | w_n, I_{n+1})$. Therefore, it remains to show that $P(q_{n+1}=True | w_n, I_{n+1}) = P(q_{n+1}=True | I_{1 \dots n+1})$.

Fix any instruction $v_t \in I_{1...{n}}$. Since sampling is thorough, there exists a recall query $q'_{n}$ for $v_t$ with a nonzero sampling probability. I.e. a query at time $n$ s.t. $P(q'_n=True | I_{1 \dots n} \setminus v_t) = 0$ and $P(q'_n=True | I_{1 \dots n}) = 1$. Fix any such query $q'_n$.

By the inductive assumption, $P(q'_n = True | w_n) = P(q'_n = True | I_{1 \dots n})$, which is equal to $1$ by definition of the recall query. But then notice that, again, by definition, the event $\{q'_n=True\}$ is equivalent to the event that the instruction $v_t$ is in the history $I_{1 \dots n}$. Therefore, $P(v_t | w_n) = 1$.

Next, we want to show that adding $I_{n+1}$ to the conditioning set of $P(v_t | w_n)$ will not change the probability. First, notice that $P(w_n, I_{n+1}) > 0$, since both $w_n$ and $I_{n+1}$ are coming from the sequence of instructions $I$ that was sampled (and hence had positive probability)\footnote{Note that we use our countability assumption here: because of it, we can have a discrete probability defined over sequences of instructions and so that any sampled instruction has positive probability}. Consequently (since $P(w_n, I_{n+1}) = P(I_{n+1} | w_n) P(w_n)$), $P(w_n, I_{n+1}) > 0$ as well. Therefore, we can condition on $w_n, I_{n+1}$ without creating a contradiction, as well as divide by $P(I_{n+1} | w_n)$. Therefore, note that $1 = P(v_t | w_n, I_{n+1}) + P(\bar{v_t} | w_n, I_{n+1})$. But $P(\bar{v_t} | w_n, I_{n+1}) = \frac{P(\bar{v_t} \cap I_{n+1} | w_n)}{P(I_{n+1} | w_n)} \leq \frac{P(\bar{v_t} | w_n)}{P(I_{n+1} | w_n)} = \frac{1 - P(v_t | w_n)}{P(I_{n+1} | w_n)} = 0$. Hence, overall, $P(v_n | w_n, I_{n+1}) = 1$.

Then, if we come back to the original query $q_{n+1}$, notice that $P(q_{n+1} | w_n, I_{n+1}) = P(q_{n+1} \cap v_n | w_n, I_{n+1}) + P(q_{n+1} \cap \bar{v_t} | w_n, I_{n+1})$, but $P(q_{n+1} \cap \bar{v_t} | w_n, I_{n+1})  \leq P(\bar{v_t} | w_n, I_{n+1}) = 0$. Thus, overall, $P(q_{n+1} | w_n, I_{n+1}) = P(q_{n+1} \cap v_t | w_n,  I_{n+1})$.

We can, therefore, proceed as follows: \begin{align*} 
 P(q_{n+1}| w_n, I_{n+1}) &= \\
 &P(q_{n+1} \cap v_t | w_n, I_{n+1}) = \\
 &P(q_{n+1} | v_t, w_n, I_{n+1}) P(v_t | w_n, I_{n+1}) = \\
 &P(q_{n+1} | v_t, w_n, I_{n+1})
\end{align*}.

We have shown that we can add any individual instruction $v_t, t \leq n$ from the conditioning set. Notice that we can repeat the reasoning with any other instruction from the history $I_{1 \dots n}$. Moreover, the reasoning holds exactly analogously if we replace $I_{n+1}$ with $I_{n+1}, v_{t_1}, v_{t_2}, \dots$, as long as all $v_{t_i}$ are instructions the true history $I_{1 \dots n}$.

 Therefore, we can add all individual instructions $v \in I_{1 \dots n}$ into the conditioning set. In other words, we get the following result: $P(q_{n+1}| w_n, I_{n+1}) = P(q_{n+1}| w_n, \{v_t: v_t \in I_{1 \dots n}\}, I_{n+1}) = P(q_{n+1} | w_n, I_{1 \dots n+1})$! 
 
 Since $w_n$ is a deterministic function from $I_{1 \dots n}$ (i.e. $w_n = \hat{\updfunc}(\hat{\updfunc}( \dots \hat{\updfunc}(\hat{\updfunc}(w_0, I_1), I_2), \dots), I_{n-1}) $, we have $P(q_{n+1} | w_n, I_{1 \dots n}) = P(q_{n+1} | I_{1 \dots {n+1}})$, which completes the proof. $\blacksquare$

\end{document}